# Generative Adversarial Networks for Generation and Classification of Physical Rehabilitation Movement Episodes

Longze Li, and Aleksandar Vakanski

University of Idaho, U.S.A.

*Abstract*— This article proposes a method for mathematical modeling of human movements related to patient exercise episodes performed during physical therapy sessions by using artificial neural networks. The generative adversarial network structure is adopted, whereby a discriminative and a generative model are trained concurrently in an adversarial manner. Different network architectures are examined, with the discriminative and generative models structured as deep subnetworks of hidden layers comprised of convolutional or recurrent computational units. The models are validated on a data set of human movements recorded with an optical motion tracker. The results demonstrate an ability of the networks for classification of new instances of motions, and for generation of motion examples that resemble the recorded motion sequences.

## I. INTRODUCTION

Patients recovering from stroke, surgery, nerves damage or bone fracture are regularly enrolled in physical therapy and rehabilitation programs to regain muscle strength, relieve pain, and improve range of motion. Both long– and short–term physical therapy provides positive results in treating musculoskeletal trauma and functional movement disorders [1], [2]. The efficiency of therapy programs is highly related to the patient adherence to prescribed exercises [3]. On the other hand, in an outpatient setting it is difficult to determine if a patient complies with the therapy program, because most of the patients do not acknowledge the incompliance [4]. The latest progress in machine learning has a potential to identify incorrect performance of exercise regimens and provide an instantaneous feedback to the patient; further, it can also provide a basis for the healthcare professionals to be proactive and take early corrective actions, if needed. The application of machine learning for evaluation of patient performance requires corresponding datasets of therapy movements for algorithm training purposes, and formulation of robust mathematical models of human body trajectories executed during physical therapy exercises.

Modeling human movements has been an essential research topic in various fields and disciplines. Congruent models of human movements furnish great benefits to ergonomic design [5], visual surveillance [6], transfer of human skills to robotic learning systems [7], etc. However, mathematical modeling of human movements remains an open research problem, due to the challenges associated with the complex stochastic and nonlinear character of the data. A current trend in machine learning related to the implementation of *deep artificial neural networks* (NNs) for modeling and representation of complex nonlinear data across various domains [8] has paved a promising path to human motion modeling.

Within the published literature on modeling human movements using machine learning approaches, most works focus on recognition and classification of movements into a particular movement type. To that end, a variety of traditional machine learning algorithms have been applied, including support vector machines, hidden Markov models, and *k*-nearest neighbors. In recent years, a body of research emerged based on the implementation of artificial NNs for the task at hand. Encoder-decoder NNs have been a commonly employed means for extraction of salient attributes in movement trajectories of captured skeletal data [9], [10]. NNs with convolutional computational units have been designed for recognition of human movements, for example, in surveillance videos [11]. Another network architecture that employs recurrent connections between the computational units has been extensively used for modeling sequential data in general [12], [13], and human motions in particular [14], [15]. Beside for movement classification task, machine learning methods have also been employed for prediction of future motion patterns, e.g., fall detection in seniors [16], or automated anticipation of driver activities [17].

Analogously, in the domain of physical therapy and rehabilitation several researchers employed machine learning for classification of patient movements [18] and for counting the number of repetitions in each exercise [19]. In [20] an intelligent robotic assistant employs machine learning for planning the next therapy session based on the patient's current progress. Similarly, machine learning-based assistants have been integrated into virtual reality therapy systems for monitoring patient performance and customizing the treatment plan according to the patient's progress [21]–[23]. In the treatment of phantom limb pain, it was found that the combination of machine learning, augmented reality, and gaming produces improved outcomes in comparison to traditional treatment approaches [24]. Another class of therapy tools employs a motion capturing camera and it displays in real-time on a screen the executed movements by the patient, and simultaneously a graphical avatar is displayed on the side of the screen that demonstrates the correctly performed movements as recommended by the physical therapist [25], [26]. These tools are excellent examples of innovative solutions and systems in support of home-based physical therapy, as they can potentially improve patient adherence to prescribed therapy programs, and subsequently, lead to reduced rehabilitation period, reduced time to functional recovery, and reduced healthcare costs.

This work presents a novel method for modeling and evaluation of physical rehabilitation exercises based on an NN architecture known as *Generative Adversarial Networks* (GANs). Introduced by Goodfellow *et al*. in 2014 [27], GAN is a deep learning model comprised of two competitive subnetworks: a generative subnetwork (commonly referred to as a generator) and a discriminative subnetwork (i.e., a discriminator). The two subnetworks are trained in an adversarial mode, where the generator improves in producing data that resemble the real input data, and the discriminator improves in distinguishing real input data from the data samples provided by the generator. GAN models have had a tremendous success in the domain of image processing, e.g., for generating super resolution photo-realistic images from text [28], face aging images in entertainment [29], blending of objects from one picture into the background of another picture, as well as in other applications, such as generating hand-written text, and music sequence generation [30].

This paper investigates the capacity of GAN models for generating human movement data related to physical therapy exercises. It was motivated by the research by Hyland *et al*. [31] where the authors designed a GAN model for generating synthetic medical data resembling the records from an intensive care unit. In general, almost all research on GANs is directed toward generating images, and only a few works have applied GANs for generating time-series data. On the other hand, the provision of means for synthesizing realistic time-series data can benefit several application areas. For the considered problem, the ability to produce movement sequences that resemble patient therapy exercises has a potential to augment the datasets of recorded therapy exercises and to lead to improved movement models. Consequently, this paper presents an evaluation of different GAN architectures for generating synthetic movement sequences. In addition, the performance of GAN networks for assessment of the level of correctness of therapy movements is also evaluated. For that purpose, soft labels are introduced for the movement repetitions based on the average deviation from a set of consistently performed movements. The study found that GANs are suitable for both generation and evaluation of therapy movement sequences.

The paper is organized as follow. Section II introduces GAN models and provides an overview of several GAN architectures relevant for the considered task. Section III describes the movement sequences data related to physical therapy exercises. The investigated architectures of the GAN models are presented in Section IV. Section V presents the validation results of using GANs for generating movement data and for evaluating exercise performance. Section VI concludes the work.

## II. INTRODUCTION TO GAN

As stated in the Introduction section, GANs consist of two subnetworks: a *discriminator D*, and a *generator G* subnetwork. The discriminator maps the input data to class probabilities, i.e., it models the probability distribution of the output labels conditioned on the input data. On the other hand, the generator models the probability distribution of the input data, which allows generating new data instances by sampling from the model distribution. Both subnetworks $D$ and $G$ are trained simultaneously in an adversarial manner, where the generator $G$ attempts to improve in creating synthetic data that approximate the input data, and the discriminator $D$ attempts to improve in differentiating the real data from the synthetically generated data.

Let's use $x$ to denote the inputs to the network, where $x \sim \mathbb{P}_r$ and $\mathbb{P}_r$ denotes the probability distribution of the real input data. The goal of the generator in GANs is to learn a model distribution $\mathbb{P}_g$ that approximates the unknown distribution of the real data $\mathbb{P}_r$. For that purpose, a random variable $z$ sampled from a fixed (e.g., uniform or Gaussian) probability distribution is used as the input to the generator, as illustrated in Fig. 1. During the training phase, the parameters of the generator are iteratively varied in order to reduce the distance, or divergence, between the distributions $\mathbb{P}_g$ and $\mathbb{P}_r$. The output of the generator is denoted $\bar{x}$ here, i.e., the generator mapping is $G: z \mapsto \bar{x}$.

To solve the described problem, a network loss function $H$ is introduced in the form of a cross-entropy,

$$H(D,G) = \mathop{\mathbb{E}}_{x \sim \mathbb{P}_r}\left[\log(D(x))\right] + \mathop{\mathbb{E}}_{\bar{x} \sim \mathbb{P}_g}\left[\log(1-D(\bar{x}))\right]. \tag{1}$$

The discriminator is trained to maximize the loss function $H$, and the generator is trained to minimize the loss function $H$, i.e.,

$$\min_G \max_D H(D,G). \tag{2}$$

In the game theory this is called a minimax game. The two subnetworks are trained in a competitive two-player scenario, where both the generator and discriminator improve their performance until a Nash equilibrium is reached. One can note that minimizing the function in (1) is equivalent to minimizing the Jensen-Shannon (JS) divergence between the real data distribution $\mathbb{P}_r$ and the model distribution $\mathbb{P}_g$.

In the case of a binary classification, the discriminator is trained to maximize $H$ by forcing $D(x)$ to approach 1 and $D(\bar{x})$ to approach 0 (Fig. 1). Contrarily, the generator is trained to minimize $H$ by forcing $D(\bar{x})$ to approach 1. Backpropagation is employed for updating the parameters of both the discriminator and generator, with the distribution $\mathbb{P}_g$ becoming more and more similar to $\mathbb{P}_r$.

The main disadvantage of GANs is the training instability. More specifically, if the generator is trained faster than the discriminator a mode collapse (also known as a Helvetica scenario) can occur, where the generator maps many values of the random variable $z$ to the same value of $x$, and reduces its capacity to learn the distribution of the real data $\mathbb{P}_r$. In addition, the model does not allow for explicit calculation of $\mathcal{P}_g(x)$, and as a result the quality of the generated data (e.g., images, as the most common data in GANs) is typically evaluated by visual observation and comparison to the actual input data. Another shortcoming of GANs is the presence of noise (and blur in the case of image data), due to the introduced random noise $z$ as input to the generator.

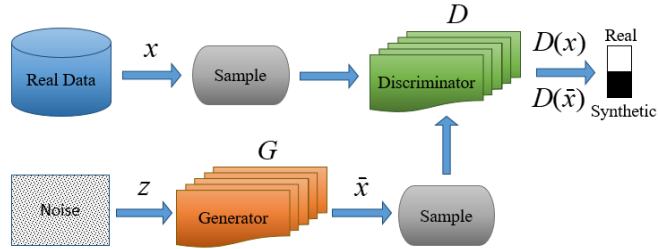

Fig. 1. A GAN model consists of a generator and a discriminator. The generator takes random noise as input and attempts to produce synthetic data that resemble the real data. The discriminator attempts to discriminate real data from the synthetic data produced by the generator.

A number of variants of GANs have been proposed since the original work, which have addressed some of the above shortcomings [32]–[35], or have been designed for domain-specific solutions [36], [37]. In the ensuing sections a brief overview of several GAN architectures is presented that are relevant for the considered problem of modeling time-series data related to patient therapy movement episodes.

*A. Deep Convolutional GAN*

Deep Convolutional GANs (DCGANs) [32] introduce several constraints and modifications to the original GAN architecture for improved stability and performance. As the name implies, the generator and discriminator subnetworks are composed of multiple layers of convolutional computational units, as opposed to the multilayer perceptron (MLP) networks proposed in the original GAN paper [27]. The modifications in DCGANs are as follows. First, the network structure in DCGANs replaces pooling layers with strided convolutions, which allows the subnetworks to adjust the spatial down-sampling and up-sampling based on the input data. Second, it eliminates fully connected layers that are commonly used after convolutional layers in deep NNs, and it relies solely on convolutional layers. Third, the DCGANs model employs batch normalization, to stabilize the gradients increase during training and reduce the possibility of a mode collapse. Batch normalization is applied to all layers, except to output layer of the generator and the input layer of the discriminator. Fourth, ReLU activation function is used for all layers in the generator, except for the last layer where a Tanh activation function is applied. For the discriminator, leaky ReLU activation function is suggested for all layers. By applying the above recommendations, the authors have demonstrated improved classification performance on various datasets of images, and capabilities of generating complex and visually realistic images.

*B. Wasserstein GAN*

Wasserstein GANs (WGANs) [33] introduce a new loss function for training the generator and discriminator subnetworks. The loss function is based on the Wasserstein distance (also known as Earth Mover distance) between the real data distribution $\mathbb{P}_r$ and the model distribution $\mathbb{P}_g$ learned by the generator,

$$W\left(\mathbb{P}_r, \mathbb{P}_g\right) = \inf_{\gamma \in \Pi(\mathbb{P}_r, \mathbb{P}_g)} \mathbb{E}_{(x,y) \sim \gamma} \left[\|x - y\|\right]. \tag{3}$$

In (3) $\Pi(\mathbb{P}_r, \mathbb{P}_g)$ denotes the set of joint distributions $\gamma(x, y)$ whose marginals are $\mathbb{P}_r$ and $\mathbb{P}_g$. In simpler terms, $\gamma(x, y)$ defines the amount of earth mass that needs to be moved from a point $x$ to a point $y$ in order $\mathbb{P}_r$ and $\mathbb{P}_g$ to be identical. Accordingly, the proposed loss function is derived as an approximation to the Wasserstein distance

$$H(D, G) = \mathbb{E}_{x \sim \mathbb{P}_r} \left[D(x)\right] - \mathbb{E}_{\bar{x} \sim \mathbb{P}_g} \left[D(\bar{x})\right]. \tag{4}$$

Such distance function induces a weaker topology than the Jensen-Shannon (JS) divergence used in the original GANs and given in (1), and the Kullback-Leibler (KL) divergence commonly used in maximum likelihood estimation. The weaker topology provides a lever for the convergence of the probability distribution of the model $\mathbb{P}_g$ to the real distribution of the data $\mathbb{P}_r$. If the discriminator $D(x)$ is a $K$- Lipschitz function, it was proven that the proposed loss function in (4) is continuous and differentiable, and produces stable gradients during training, thereby improving the problem of training instability in GANs.

In addition, the values of the adopted loss function $H$ in (4) are correlated to the quality of the generated data samples by the generator, and with that WGANs provide a basis for quantifying the performance of the generator, rather than relying on visual observation of the generated samples. Accordingly, during the network training, the loss function is used to evaluate the training convergence, i.e., to identify if the network is being trained.

To enforce a Lipschitz constraint on the discriminator, it was proposed to apply clipping of the parameters into a range $[-c, +c]$ after each gradient update, where $c$ is a referred to as a clipping constant. The suggested value for $c$ in the paper is 0.01.

Unlike GANs, the output of the discriminator in WGANs is not a probability; instead, it is an estimate of the Wasserstein distance between the distributions. Therefore, the authors use the term critic in the article, rather than discriminator, due to the similarity with the actor-critic methods in reinforcement learning.

*C. Recurrent GAN*

Recurrent GAN (RGAN) [31] is an alternative GAN model that is designed for handling multi-dimensional time-series data. For that purpose, recurrent computational units are employed for the discriminator and generator. More specifically, a layer of unidirectional Long Short-Term Memory (LSTM) computational units [12] is used for both subnetworks.

The proposed approach was applied to medical records data from an intensive care unit. The authors investigated the ability of RGAN to generate synthetic medical data samples and the potential for use in data augmentation in cases of insufficiency of real data for training deep learning models. In the article, RGAN was also implemented for processing synthetic sine waves sequences, as well as images. The authors claim that RGAN is more suitable for dealing with time-series data in comparison to the proposed GAN alternatives composed of layers of convolutional kernels.

III. DATA

*A. Data Description*

The presented GAN models are validated on the University of Idaho – Physical Rehabilitation Movements Data (UI–PRMD) set. The full description of the dataset is provided in [38], and here only the most relevant data details are presented. The motion sequences related to two common training movements in physical therapy exercises—a deep squat, hereafter Movement 1, and a standing shoulder abduction, hereafter Movement 2—are used in this work. A Vicon optical tracking system was used for the data collection, which employs eight high-resolution cameras for tracking the position of 39 reflective markers attached to strategic locations on a subject's body. The optical tracking system captured the executed motions at 100 frames per second, while a dedicated software program assembled the recorded data into sequences of joint angle positions. The output data by the motion capture system are time-series consisting of 117–dimensional vectors of joint angle displacements.

Ten healthy subjects performed 10 repetitions for each of the two movements. In addition, the subjects performed 10 repetitions for each movement in an incorrect fashion, simulating performance by patients with musculoskeletal constraints that preclude them from executing the movements in a manner prescribed by the physical therapist. The single repetitions of each movement were separated, by identifying the beginning and end time steps of each repetition. Consequently, this resulted in a dataset consisting of 100 instances of correctly performed repetitions, and 100 instances of incorrectly performed repetitions, for each movement. By elimination of poorly recorded repetitions, as well as elimination of the data of subjects who performed the standing shoulder abduction exercise with their left arm (versus the rest of the subject who used their right arm), the final number of repetitions was reduced to 90 samples for Movement 1, and 63 samples for Movement 2. The number of correct and incorrect repetitions was kept equivalent for the two movements.

*B. Data Notation*

The number of repetitions of a movement is denoted $N$, and the sequence of measurements by the optical tracking system for each correctly performed repetition is denoted $\mathbf{U}_n$, where $n$ is used to index the individual sequences. The set of correct repetitions of a movement forms $\mathcal{U} = \{\mathbf{U}_n\}_{n=1}^{N}$. Each sequence $\mathbf{U}_n$ contains $M$ temporally ordered vectors $\mathbf{U}_n = \left(\mathbf{u}_n^{(1)}, \mathbf{u}_n^{(2)}, \ldots, \mathbf{u}_n^{(M)}\right)$, where each temporal measurement is a $D$-dimensional vector, i.e., $\mathbf{u}_n^{(m)} \in \mathbb{R}^D$. The adopted notation employs bold fonts for vectors and matrices.

Similarly, the set of incorrect repetitions of the movements is denoted $\mathcal{W} = \{\mathbf{W}_n\}_{n=1}^{N}$. Each movement sequence $\mathbf{W}_n$ consists of $M$ vectors $\mathbf{w}_n^{(m)} \in \mathbb{R}^D$, for $m = 1, 2, \ldots, M$.

*C. Data Preprocessing and Labeling*

The data preprocessing included scaling of the angular displacement measurements in the range $[-1, +1]$. More specifically, all sequences in the correct and incorrect movement sets were divided by the maximum absolute value of the correct set, i.e., $\max\left(\left|u_n^{(m)}\right|\right)$ for $n = 1, 2, \ldots, N$, $m = 1, 2, \ldots, M$. In addition, each movement sequence $\mathbf{U}_n$ and $\mathbf{W}_n$ was zero-mean shifted. Although it is commonly recommended to normalize the inputs to NNs into data vectors with a variance of 1, this is not applicable to the movement data since the variability of the individual dimensions is an important attribute of the data and needs to be preserved.

As the goal of the considered task is to evaluate the level of correctness in the execution of movement repetitions during rehabilitation exercises, soft labels are assigned to each repetition instance. Root-mean-squared (RMS) deviation was adopted here as a metric for assessment of the repetition consistency. For this purpose, the RMS distance between each correct sequence $\mathbf{U}_n$ and the entire set $\mathcal{U}$ is calculated, i.e.,

$$\xi_i = \frac{1}{N} \sum_{n=1}^{N} \sqrt{\frac{1}{M} \sum_{i=1}^{M} \left(\mathbf{u}_n^{(i)} - \mathbf{u}_n^{(m)}\right)^2}, \text{ for } i = 1, 2, \ldots, N. \quad (5)$$

Similarly, the RMS distance between each incorrect repetition $\mathbf{W}_n$ and the set of correct movements $\mathcal{U}$ is calculated as

$$\zeta_i = \frac{1}{N} \sum_{n=1}^{N} \sqrt{\frac{1}{M} \sum_{i=1}^{M} \left(\mathbf{w}_n^{(i)} - \mathbf{u}_n^{(m)}\right)^2}, \text{ for } i = 1, 2, \ldots, N. \quad (6)$$

One can note that in (6) the RMS deviation is calculated with respect to the set of correct movements.
Soft labels are assigned next to each of the correct and incorrect data sequences as follows:

$$l_i = \frac{1 - \xi_i - \overline{\xi}}{\tau}, \; l_i = \frac{1 - \zeta_i - \overline{\xi}}{\tau}, \text{ for } i = 1, 2, \ldots, N. \quad (7)$$

The resulting soft labels for the two movements are shown in Fig. 2. The parameter $\tau$ in (7) is a normalization factor that was empirically assigned the value of 100 for Movement 1 and 200 for Movement 2. The labels in (7) were set with a goal to be distributed in the range $[0, +1]$, and to retain a separation boundary between the correct and incorrect movements. It can be noticed in Fig. 2 that several of the correct movements are performed in an inconsistent manner, and they are less similar to the remaining correct set of movements than some of the incorrectly performed movements. That was one motivation to introduce soft labels for the movement instances, instead of employing hard labels of 1's for the correct movements and 0's for the incorrect movements.

Furthermore, as stated earlier one of our objectives is to assess the potential of GAN models for evaluation of the level of correctness of therapy movements. The provision of soft labels allows to train an NN on a set of correct and incorrect movements, and to validate the trained networks on another set of correct and incorrect movements. Also, with the use of soft labels the problem was cast from binary classification into a one-class classification, where all data instances belong to the same class of movement but have varying levels of movement quality. In addition, we believe that the use of soft labels provides richer information of the input data and a basis for an improved performance of both the generator and discriminator subnetworks.

One final note regarding the above procedure for applying soft labels to the motion data is that RMS deviation is probably a suboptimal metric for quantifying the distance between the high-dimensional data sequences. Although it was adopted here for proof of concept, the selection of metrics for the task at hand is one of the authors' topics for future research.

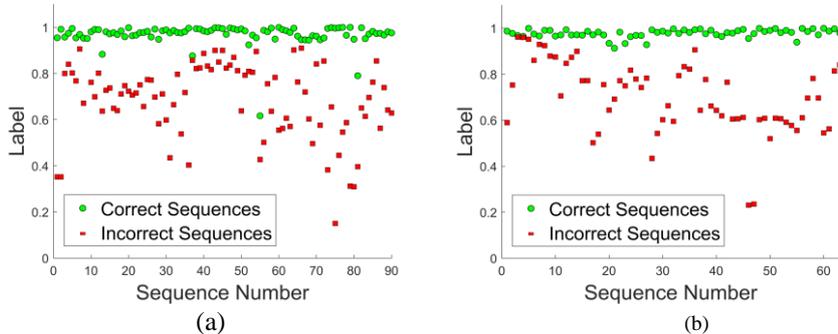

Fig. 2. Soft labels for: (a) Deep squat movement; (b) Standing shoulder abduction movement. The labels for both correct and incorrect sequences for the movements are shown in the figure.

## IV. NETWORK ARCHITECTURES

The paper investigates the GAN variations presented in Section II (and their sub-variants in one case). A basis for comparison of the considered architectures is the DCGAN model depicted in Fig. 3. The generative subnetwork consists of one fully connected layer and three padded convolutional layers. Following the guidelines in the DCGAN paper [26], ReLU activation functions are used in the generator except in the last layer that uses Tanh activation, and strides are utilized instead of pooling layers. As illustrated in Fig. 3, the discriminative subnetwork has three padded convolutional layers. Leaky ReLU activation functions are introduced in the discriminator, and a dropout rate of 20% was applied to prevent overfitting. Adam optimizer was the choice in both subnetworks.

The investigated GAN models are fully described in Table I. The networks' structures are based on the DCGAN model presented in Fig. 3. The networks are explained in more detail in the next section.

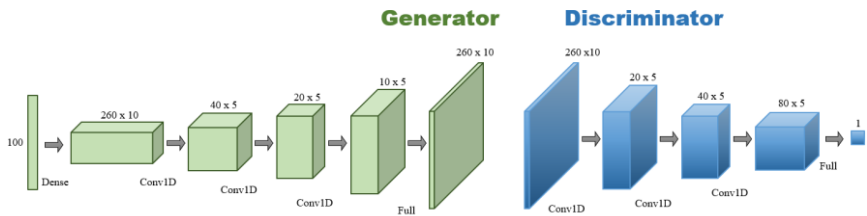

Fig. 3. DCGAN model layers consisting of a generator and discriminator subnetworks composed of convolutional and MLP layers of hidden computational units.

TABLE I: GAN NETWORK ARCHITECTURES[1]

| Network | Generator | Discriminator |
|---|---|---|
| GAN | 50 (LR) × 100 (LR) × 200 (LR) × $M$ = 260, $D$ = 10 (TH): Adam | 100 (LR,D) × 50 (LR,D) × 1 (S): Adam |
| DCGAN-1 | 100 (R, BN) × $M$ = 260, $D$ = 10 (R, BN) × Conv1D (40, 5, R, BN) × US(2) × Conv1D (20, 5, R, BN) × US(2) × Conv1D ($D$ = 10, 5, TH): Adam | Conv1D (20, 5, LR, D, St:2) × Conv1D (40, 5, LR, D, BN) × Conv1D (80, 5, LR,D, BN) × 1 (S): Adam |
| DCGAN-2 | 100 (LR, BN) × $M$ =260, $D$ = 10 (LR) × Conv1D (40, 5, LR) × US(2) × Conv1D (20, 5, TH) × US(2) × Conv1D ($D$ = 10, 5, TH): Adam | Conv1D (10, 5, LR, D, St:2) × Conv1D (20, 5, LR, D) × Conv1D (40, 5, LR,D) × 50 (LR,D) × 1 (S): Adam |
| WGAN | 100 (LR) × $M$ = 260, $D$ = 10 (LR) × Conv1D (40, 5, LR) × US(2) × Conv1D (20, 5, LR) × US(2) × Conv1D ($D$ = 10, 5, TH): Adam | Conv1D (10, 5, LR, D, St:2) × Conv1D (20, 5, LR, D) × Conv1D (40, 5, LR,D) × 50 (LR,D) × 1 (S): SGD |
| RGAN | ($M$ = 260,5) × LSTM(100) : Adam | LSTM(100) × 1 (S): SGD |

[1] Acronyms: LR – Leaky ReLU activation, R – ReLU activation, TH – Tanh activation, S – Sigmoid activation, BN – Batch normalization, US – Upsampling, D – Dropout, St – Strides, SGD – Stochastic Gradient Descent.

## V. RESULTS

*A. Movement Generation*

The performance of the GAN representations listed in Table I is examined in relation to their capacity to generate data samples that resemble the time-series data of the actual physical therapy movements.

A subset of the data with reduced dimensionality is first considered, where 10 dimensions with the largest variation are extracted and used as input to the network. Several examples of the sequences for Movement 1 are presented in Fig. 4(a).

One undesirable effect in the synthetic data samples produced by the GAN models is the distortion of the ends and beginnings of the generated sequences. To reduce the effect of the distortions, 10 time steps of synthetic data were added at the beginning and at the end of each sequence. The beginning 10 time steps are set equal to the first vector in each sequence, and the ending 10 time steps are set equal to the last vector in the sequence. Consequently, for Movement 1 the number of time steps $M$ was increased from 240 to 260, and for Movement 2 the length $M$ was increased from 231 to 251 time steps.

The GAN architectures in Table I are related to processing the input data for Movement 1, with the number of time steps $M = 260$, and dimensionality $D = 10$. The NNs for Movement 2 and for the presented cases with different dimensionality have the same structure as the GANs presented in Table I, and only the parameters $M$ and $D$ are varied.

For Movement 1, the subset for training purposes includes 70 correct and 70 incorrect movement repetitions, and the validation subset consists of the remaining 20 correct and 20 incorrect sequences. Similarly, for Movement 2, the training and validation subsets have 98 and 28 sequences of correct and incorrect repetitions, respectively.

The sequences generated with the original GAN model based on the structure outlined in Table I and consisting of MLP layers of computational units are shown in Fig. 4(b). Conclusively, the data is quite noisy, and the network experiences a mode collapse early in the training, failing to refine the output of the generator. The next examined model is DCGAN-1 from Table I, which implements the network structure recommended by the authors of [26]. However, the model was not able to produce data that resemble the real motion sequences. One potential reason is that the DCGAN network design reported in [26] is more applicable to image data. The suggested batch normalization of the hidden layers was the main contributing factor for the network failure with the human movement input data. Nevertheless, a variant of the model listed in Table I as DCGAN-2 provided realistic synthetic data. This network employs convolutional layers of units in a slightly altered architecture in comparison to the recommended DCGAN-1 model. Several representative examples of the generated sequences by DCGAN-2 are shown in Fig. 4(c). Next, instances of the synthetic data generated with WGAN are shown in Fig. 4(d). The quality of the data is comparable to the sequences generated with DCGAN-2. Overall, WGAN model exhibited improved stability during training and, to a certain extent, visually improved quality of generated data. The last investigated model is RGAN with the network structure presented in Table I, consisting of recurrent LSTM computational units. A set of generated data is displayed in Fig. 4(e). The RGAN model created the smoothest synthetic sequences for Movement 1, and it outperformed the other models that are based on convolutional and MLP layers of hidden units.

Another validation case is presented next for Movement 2, related to the standing shoulder abduction exercise. In this case, the time-series dimensionality is reduced to the three dimensions with the largest variance. Considering the strong correlation between the joint angular displacements in human movements, a body of work in the literature relies on only several most important dimensions for motion modeling. As expected, for the considered motion the dimensions with the largest variability correspond to the angular displacements of the upper arm, lower hand, and the wrist. Two movement repetitions as acquired by the optical tracker are displayed in Fig. 5(a). Similar to the first validation case, the networks presented in Table I are employed for modeling the movements and generating synthetic data samples. Instances of the generated sequences with the conventional GAN model are shown in Fig. 5(b), and similar to Fig. 4(b), the sequences are quite noisy. Examples of the generated data with the DCGAN-2 and WGAN models are shown on Figs. 5(c) and (d), respectively. The quality of the GAN-generated sequences is visually appealing, and one can notice that the networks demonstrated improved performance in the case of low-dimensional input data. Conversely, the samples generated with DCGAN-2 are less smooth for this movement. The generated data with RGAN is presented in Fig. 5(e).

In summary, the RGAN model produced the smoothest and visually attractive synthetic movement sequences for the two movements. The GAN models based on layers of convolutional kernels were also able to generate data sequences of comparable and acceptable quality. The synthetic data samples produced with the original GAN model are the least smooth when compared to the other cases, although the model was able to learn the general pattern of the movement sequences.

*B. Movement Classification*

Next, the ability of the GANs presented in Table I to classify therapy movement repetitions is evaluated. For comparing the performance of the models, a metric is adopted which sums the absolute differences between the predicted probabilities of the discriminator and the soft labels for the data instances $\mathbf{X}_k$ in the validation subset, i.e.,

$$C = \sum_{k=1}^{K} |\mathcal{P}(\mathbf{X}_k) - l_k|, \tag{8}$$

where $K$ denotes the number of validation sequences.

The values of the metric $C$ for the considered GAN models are presented in Table II. Presented in the table also are the performance scores of NNs consisting only of the discriminator subnetwork (i.e., without a generator subnetwork). In Table II, the corresponding NNs have an extension "-Disc." The values of the metric for the WGAN model are not presented in the table, as the outputs of its discriminator are not probabilities (but are values of the Wasserstein distance). Table II contains the distances $C$ for cases of 3-dimensional and 10-dimensional movement sequences.

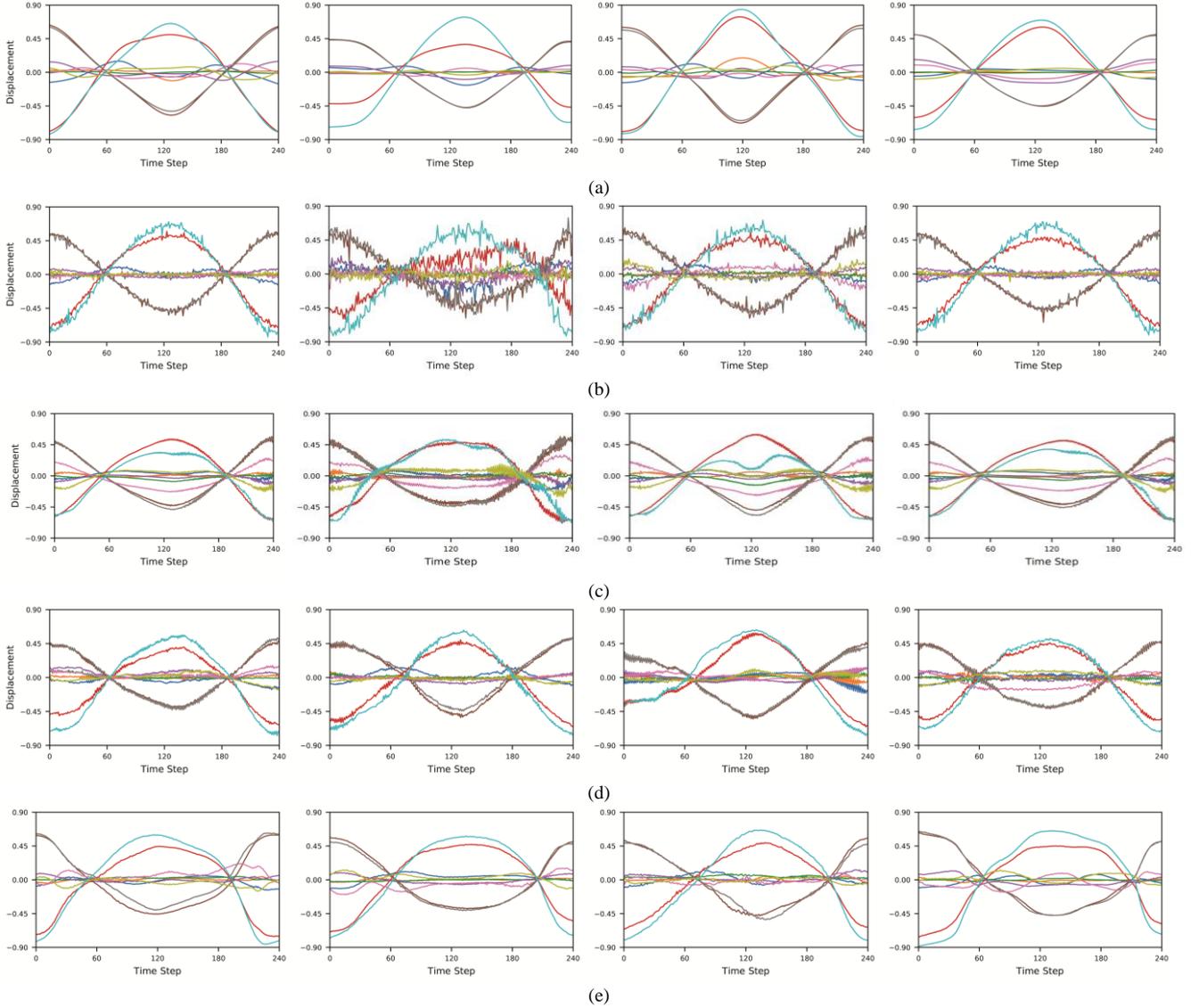

Fig. 4. (a) Samples of 10-dimensional Movement 1 sequences as recorded with the optical tracking system. (b) Examples of generated sequences with the GAN network from Table I. (c) Examples of generated sequences with the DCGAN-2 network from Table I. (d) Examples of generated sequences with the WGAN network from Table I. (e) Examples of generated sequences with the RGAN network from Table I.

For the discriminative NNs in Table II, the presented numbers correspond to the average value of the parameter $C$ based on five runs of the models. The values in the parenthesis preceded with the symbol S are the respective standard deviations. Early stopping of 100 epochs was employed in the training phase.

For the GAN models, the presented values of the parameter $C$ in Table II are based on a single run of the networks. In particular, the values in the parenthesis preceded with the symbol M are the minimum values of the parameter $C$, whereas the upper numbers represent the average values of the parameter $C$ based on the preceding 25 epochs and the succeeding 25 epochs relative to the minimum value. Averaging was employed in order to filter out the significant oscillations in the obtained $C$ values with the GAN models.

One example of the performance of the considered models is depicted in Fig. 6. The figure shows the soft labels calculated based on (7) and the output probabilities of the DCGAN-1-Disc model. Fig. 6(a) displays the scores for Movement 1, which has a validation set of 40 sequences. In the figure, the first 20 sequences are drawn from the set of correct movements, and the last 20 sequences are drawn from the set of incorrect movements. One can notice that the network evaluates the correct movements very accurately, and that for the incorrect movements the network predictions are close to the assigned labels. Similarly, Fig. 6(b) presents the labels and the network predictions for Movement 2, for which the validation set consists of 28 data sequences. The predicted labels for the movement repetitions for this case also approximate the actual labels.

From the results in Table II regarding the discriminative NNs, it can be concluded that DCGAN-2-Disc achieved the lowest cumulative deviation between the input soft labels and the predicted labels, in comparison to the other discriminative models. Overall, the 10-dimensional sequences provided richer discriminative information of the movements and produced better results in comparison to the 3-dimensional sequences. The discriminators of the original GAN and DCGAN-1 also achieved comparable classification accuracy.

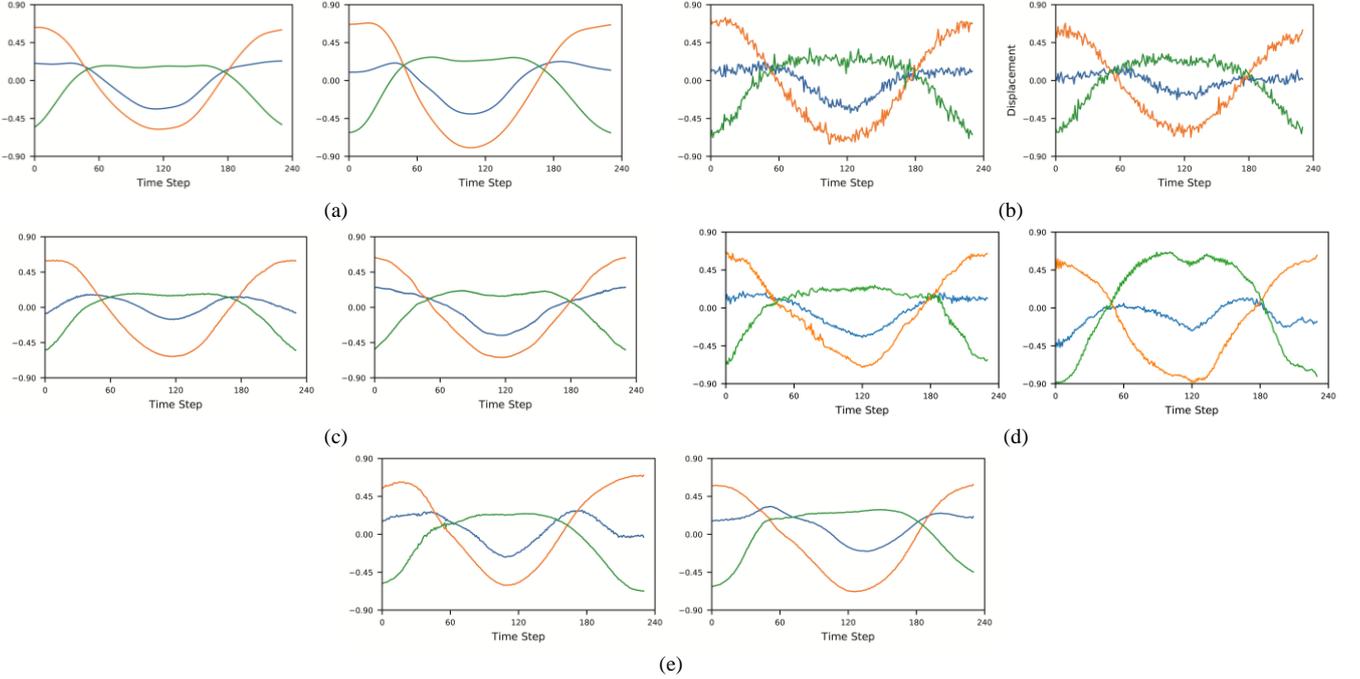

Fig. 5. (a) Samples of 3-dimensional Movement 2 sequences as recorded with the optical tracking system. (b) Examples of generated sequences with the GAN network from Table I. (c) Examples of generated sequences with the DCGAN-2 network from Table I. (d) Examples of generated sequences with the WGAN network from Table I. (e) Examples of generated sequences with the RGAN network from Table I.

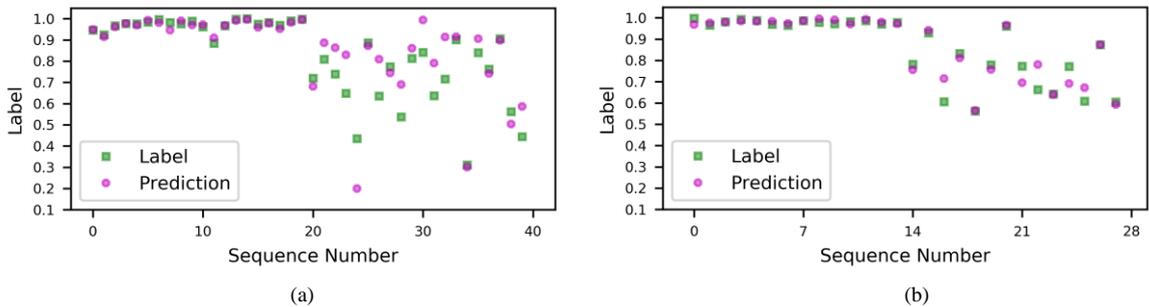

Fig. 6. Soft labels and predicted labels by the DCGAN-1-Disc model for: (a) Deep squat movement; (b) Standing shoulder abduction movement.

In comparison to the discriminative NNs, the predicted labels of the movements by the GAN architectures are characterized by lower deviation values $C$ in relation to the input labels. The obtained values are shown with a bold font in Table II. Almost in all cases, the GAN models outperformed the discriminative NNs. The discriminator based on recurrent computational units RGAN-Disc produced lower or comparable classification accuracies, compared to the models with convolutional units. The RGAN demonstrated lower classification accuracy and the results are not shown in the table.

Among the drawbacks of employing GANs for this task is the computational expense, as the GAN networks took significantly longer to train in comparison to the discriminative NNs, and in some cases, the GAN models required an additional fine-tuning of the hyperparameters to obtain the reported classification accuracy.

TABLE II: Classification Accuracy Results for the Considered GAN Models and the Corresponding Discriminative Models[1]

| Network | Movement 1 | | Movement 2 | |
|---|---|---|---|---|
| | 3D | 10D | 3D | 10D |
| GAN | 2.220 (**M1.821**) | 2.097 (**M1.791**) | 0.801 (**M0.582**) | 0.797 (**M0.607**) |
| GAN-Disc | 2.254 (S±0.053) | 2.683 (S±0.145) | 1.008 (S±0.101) | 0.922 (S±0.042) |
| DCGAN-1 | 3.965 (**M2.601**) | 2.237 (**M2.001**) | 1.136 (M0.989) | 0.789 (**M0.614**) |
| DCGAN-1-Disc | 3.251 (S±0.637) | 2.413 (S±0.058) | 0.866 (S±0.025) | 0.852 (S±0.225) |
| DCGAN-2 | 3.649 (**M1.865**) | 1.999 (**M1.336**) | 0.836 (**M0.745**) | 0.793 (**M0.645**) |
| DCGAN-2-Disc | 2.309 (S±0.160) | 2.057 (S±0.318) | 0.799 (S±0.016) | 0.947 (S±0.005) |
| RGAN-Disc | 2.637 (S±0.160) | 2.446 (S±0.455) | 1.336 (S±0.149) | 0.878 (S±0.046) |

[1] M – minimum value; S – standard deviation.

## VI. Conclusion

The article employs GANs for modeling and evaluation of physical rehabilitation movements. Four relevant GAN models are considered, which include: GAN, DCGAN, WGAN, and RGAN. The ability of the networks to generate data instances that resemble two sets of therapy movements is evaluated. Further, the classification accuracy of the GANs is assessed based on introduced soft labels for the movement sequences. The presented results demonstrate the capacity of the considered GAN models to learn the underlying structure of the movement sequences, and with that, to generate realistic synthetic movement data, and to predict the level of performance consistency on a set of unseen movement sequences. These capabilities furnish a potential for augmentation of datasets of therapy movements with synthetically generated samples for improved movement modeling, and for utilization in automated monitoring and evaluation of the level of correctness of patient movements in home-based therapy programs.

## VII. Acknowledgement

This work was supported by the Center for Modeling Complex Interactions through NIH Award #P20GM104420 with additional support from the University of Idaho.